\documentclass[12pt]{article}

\usepackage{amsmath,amsthm, amsfonts, amssymb, amsxtra, amsopn}
\usepackage{pgfplots}
\usepgfplotslibrary{colorbrewer}
\pgfplotsset{compat = 1.15, 
			 cycle list/Set1-3} 
\usetikzlibrary{pgfplots.statistics, pgfplots.colorbrewer} 
\usepackage{pgfplotstable}
\usepackage{graphicx,grffile}
\usepackage{multirow}
\usepackage{booktabs}
\usepackage{tcolorbox}
\usepackage{algorithm} 
\usepackage[noend]{algpseudocode} 
\usepackage{listings}
\usepackage{cmap}
\usepackage{colortbl}
\usepackage{adjustbox}
\usepackage{epsfig}

\usepackage[tableposition=top,font=small,skip=5pt,width=\textwidth]{caption}
\usepackage{subcaption}
\usepackage{makecell}

\usepackage[explicit]{titlesec}

\usetikzlibrary{patterns}

\def\bbb#1{{\color{blue}#1}}
\def\com#1{\bbb{\texttt{/\kern-1.5pt /} #1}}

\def\tau{\mathcal{T}}

\PassOptionsToPackage{hyphens}{url}
\PassOptionsToPackage{table}{xcolor}

\usepackage{hyperref}
\hypersetup{colorlinks=true,linkcolor=black,citecolor=black,urlcolor=blue,filecolor=black}
\hypersetup{pdfpagemode=UseNone,pdfstartview=}

\definecolor{darkgreen}{rgb}{0.125,0.5,0.169}

\usepackage[shortlabels]{enumitem}
\setlist[itemize]{noitemsep, topsep=0pt}

\advance\oddsidemargin by -0.45in
\advance\textwidth by 0.9in

\advance\topmargin by -0.5in
\advance\textheight by 1.0in

\long\def\symbolfootnotetext[#1]#2{\begingroup%
  \def\thefootnote{\fnsymbol{footnote}}\footnotetext[#1]{#2}\endgroup}

\newcommand\dunderline[3][-1pt]{{%
      \sbox0{#3}%
      \ooalign{\copy0\cr\rule[\dimexpr#1-#2\relax]{\wd0}{#2}}}}
\def\uuu{\kern-1pt\dunderline{0.75pt}{\phantom{M}}}

\usetikzlibrary{positioning}
\usepgfplotslibrary{colorbrewer,groupplots}

\title{CAM-Guided Saliency Cutout and Image-Based Malware Classification}

\author{Yasaman Ebrahimi\footnotemark[1]\,\,\footnotemark[3]\ \ \ 
Martin  Jure\v{c}ek\footnotemark[2]\ \ \
Mark Stamp\footnotemark[1]\,\,\footnotemark[4]} 

\begin{document}

\symbolfootnotetext[1]{Department of Computer Science, San Jose State University}
\symbolfootnotetext[2]{Faculty of Information Technology, Czech Technical University in Prague}
\symbolfootnotetext[3]{yasi.ebrahimi0$@$gmail.com}
\symbolfootnotetext[4]{mark.stamp$@$sjsu.edu}

\maketitle

\abstract
Dropout regularization is commonly used to reduce overfitting by removing parts of a neural network during training. For Convolutional Neural Networks (CNN), cutouts serve a somewhat analogous purpose. Cutouts can be implemented as data augmentation: the original training image is retained, and additional copies are created with regions removed. In this chapter, we test whether cutout placement can be improved by using High-Resolution Class Activation Mapping (HiResCAM). We compare four controlled training conditions: no cutout, standard random cutout, low-saliency cutout, and high-saliency cutout. We experiment using grayscale malware images from the RawMal-TF dataset (17 families with~1,000 samples per family), and for comparison to natural images, we experiment with the well-known CIFAR-100 dataset. All experiments are based on ResNet18 with~100 training epochs. For the cutout experiments, we test cutout areas of~5\%, 10\%, 20\%, and~30\%, and we consider~$M\in\{4,8\}$ augmented copies per original training image. The RawMal-TF results are slightly worse for all three cutout cases (random, high and low saliency) as compared to no cutouts. In contrast, our CIFAR-100 experimental results improve slightly under low-saliency cutout. These results suggest that the value of saliency-guided cutout is domain dependent, and that malware images should not be treated as equivalent to natural images.

\bigskip

\noindent\textbf{Keywords}: HiResCAM $\cdot$ ResNet18 $\cdot$ RawMal-TF $\cdot$ CIFAR-100 $\cdot$ Cutout regularization
$\cdot$ Malware classification $\cdot$ Convolutional Neural Networks

\section{Introduction}

Image-based malware classification is a process that turns binary files into images so that computer vision models
can be used to analyze malware. Convolutional Neural Network (CNN) models often achieve state-of-the-art results
on challenging malware classification tasks. The main goal of this research is to evaluate whether cutout regularization
guided by High-Resolution Class Activation Mapping (HiResCAM) improves image-based malware classification,
and to compare this behavior with analogous results on natural-image datasets.

Standard cutout removes a randomly selected region from an image during training, which
encourages a CNN to avoid relying too heavily on a single local region~\cite{cutout}. In this work, we keep
the structure of standard cutout but change the rule used to choose the cutout location. Instead of selecting
a square cutout region randomly, we use HiResCAM saliency maps from a trained teacher model to choose
either a low-saliency region or a high-saliency region. The goal is to determine whether saliency can guide 
a cutout-style data augmentation method in a way that improves classification results.

The distinction between permanent masking and cutout augmentation is important. A permanent saliency mask tests
whether the image can be simplified by deleting regions. In contrast, a cutout regularizer tests whether a model becomes
more stronger when it is forced, during training, to classify partially occluded copies of the original images. These are
different experimental questions. The present study therefore compares saliency-guided cutout directly against
standard random cutout and a no-cutout baseline. This makes random cutout the primary control condition, since
it uses the same number of augmented copies and the same cutout area, but does not use saliency to select the
square cutout region.

Our main malware dataset is a collection of malware images derived from the
RawMal-TF dataset~\cite{stamp2024malwareimages_arxiv,rawmaltf}. RawMal-TF is ideal for this study because
it contains a large collection of malware samples labeled by family, and because each sample can be represented
through multiple image transformations. Based on the goal of a controlled experiment, this chapter limits the
RawMal-TF study to one image type, namely, standard grayscale images. 
For comparison, we conduct analogous experiments with the CIFAR-100~\cite{cifar100} natural-image dataset.

The research questions addressed in this chapter are the following:
\begin{itemize}
\item Does saliency-guided cutout improve malware family classification based on grayscale images, 
as compared with no cutout?
\item Does saliency-guided cutout improve over standard random cutout when the number of cutout copies
and cutout area are held constant?
\item Is low-saliency cutout or high-saliency cutout more effective for malware image classification?
\item Are the conclusions robust across multiple seed values and cutout areas?
\item Is the behavior observed on malware images consistent with the behavior 
observed on natural images?
\end{itemize}

The main finding is negative for the malware task. That is, for the grayscale RawMal-TF images,
the no-cutout ResNet18 baseline attains the best mean validation accuracy among all tested conditions. Standard
random cutout, low-saliency cutout, and high-saliency cutout all reduce mean validation accuracy on this malware dataset.
The expanded seed and cutout-area sweeps show that low-saliency cutout is sometimes slightly better than random
cutout and sometimes slightly worse, while high-saliency cutout is generally worse than random cutout. This result
is interesting, because it shows that cutouts do not simply transfer from generic image regularization to malware
images. The analogous CIFAR-100 results provide a useful contrast: low-saliency cutout improves over no cutout for several
settings, while high-saliency cutout is consistently poor. This contrast suggests that saliency-guided cutout is
potentially useful, but its usefulness depends on the image domain. These results also highlight that malware
images are distinct from natural images, and hence techniques developed for natural images may not carry over
to the field fo malware analysis.
 
The remainder of this chapter is organized as follows. Section~\ref{sect:back} covers relevant background topics,
including related work. In Section~\ref{sect:meth} we outline the methodology used in subsequent experiments.
Our experimental results are given in Section~\ref{sect:res}. Section~\ref{sect:dis} discusses the main findings,
limitations, and implications for our malware-image experiments, along with suggestions for future work. 
In Section~\ref{sect:con}, we conclude the chapter with a summary of our main results.

\section{Background and Related Work}\label{sect:back}

This section reviews the main ideas needed to interpret our experiments. We first discuss dropout, cutout,
and saliency-guided masking, as related to regularization and masking strategies. We then describe image-based
malware classification and discuss why CAM-guided masking might be expected to behave differently for malware 
images than for natural images.

\subsection{Dropout, Cutout, and Data Augmentation}\label{sect:DCDA}

Dropout regularization is a common method used to reduce overfitting during neural network training. It works by
randomly removing hidden and input units during training, which discourages the model from depending too much on specific
features and can improve generalization~\cite{dropout}. Cutout applies a somewhat similar idea to image data by masking random
spatial regions of the input image during training~\cite{cutout}. For convolutional neural networks, this can help the model
learn from different visual patterns instead of focusing too much on one region.

The standard cutout setting is most naturally viewed as data augmentation. A training image is retained in its original form,
and additional training copies are created, typically with one square region removed. The model is then trained on both
the original and occluded images. Note that validation and test images are not
occluded, as the purpose is to improve a model that will later be evaluated on the original distribution. 
Therefore, a cutout method is successful only if it improves validation accuracy on the
original images, not merely if the model learns to classify artificially masked images.

The experiments in this chapter use a simple and controlled version of this idea. When cutout is enabled, each original
training sample contributes the original image plus~$M$ augmented cutout copies. The values~$M=4$ and~$M=8$
are tested. For all cutout conditions, cutout areas of~5\%, 10\%, 20\%, and~30\%\ of the image area are evaluated.
The random, low-saliency, and high-saliency settings differ in how the cutout region is selected, which is the
core comparison needed for our research problem.

\subsection{CAM and HiResCAM}

Instead of masking random image areas, we propose a Class Activation Map (CAM) based method that can be used
to identify important regions. Grad-CAM produces a class-discriminative heatmap by using gradients from a target
convolutional layer to identify image regions that contribute to a prediction~\cite{selvaraju2017gradcam}. HiResCAM
follows the same general approach, but it is designed to provide more faithful explanations by avoiding the
gradient-averaging step used in Grad-CAM~\cite{hirescam2020}. For a selected class, HiResCAM records the
activations from a target convolutional layer and computes the gradient of the class score with respect to those
activations. The activation at each spatial location is multiplied by its corresponding gradient, and these activation-gradient
products are summed across the feature-map channels. The resulting heatmap is then passed through a ReLU operation,
normalized, and resized to match the input image size. Larger heatmap values indicate regions that contribute more
strongly to the model's prediction, while smaller values indicate regions that contribute less strongly.

A recent malware-visualization study used CAM methods, including HiResCAM, to interpret malware image classifiers and to
guide masking strategies~\cite{brosolo2025static}. This is part of a broader explainable-malware-analysis literature in which
saliency maps and post hoc explanations are used to inspect classifier behavior~\cite{jei2023saliency,xai,grad_cam}.
Other saliency and perturbation-based approaches, such as RISE and deletion/insertion tests, further show that the choice
of explanation method can affect how image evidence is interpreted~\cite{hama2023deletion,petsiuk2018rise}. 
That work motivates the idea that CAM maps may contain useful
information about which image regions matter for malware classification. However, the present work uses CAM maps
for a different purpose. We are not testing whether removing low-saliency pixels produces a cleaner image;
instead, we are testing whether CAM maps can guide cutout augmentation in a way that is analogous to standard
random cutout.

A crucial distinction in this study is between low-saliency cutout and high-saliency cutout. Low-saliency cutout removes
a square region selected from areas that the teacher model treats as less important. This tests whether the model benefits
from occlusions placed in regions that are expected to be less informative. High-saliency cutout removes a square region
selected from areas that the teacher model relies on most strongly. This is closer to the regularization intuition behind cutout:
hiding highly informative regions may force the student model to learn from other available features. Although both strategies
come from the same HiResCAM heatmap, they represent different research questions and should not be construed as the
same type of masking.

\subsection{Image-Based Malware Classification}

Malware image classification treats executable bytes or extracted binary features as image-like structures that can be
used by computer vision models. Early work on malware visualization showed that malware binaries can be converted
into grayscale images and classified using visual patterns~\cite{nataraj2011malware}. Subsequent work has considered
handcrafted image descriptors, transfer learning, convolutional neural networks, and other deep learning approaches for
image-based malware classification~\cite{Bhodia19,jain,Stamp2021,Yajamanam18}. More recent work has shown
that different binary-to-image transformations can produce different visual representations of the same malware sample.
Note that there is no universally accepted ``best'' image transformation for malware
classification~\cite{stamp2024malwareimages,stamp2024malwareimages_arxiv}.

The main malware dataset used in this study is RawMal-TF, which contains malware samples labeled by type
and family~\cite{rawmaltf}. The malware images used in our experiments consist of~17 malware families from
RawMal-TF, with~1,000 malware samples per family. For each malware sample, eight distinct image representations
are available, giving~17,000 malware samples and~136,000 total images~\cite{stamp2024malwareimages_arxiv}.
In the present experiments, we do not consider all eight image representations. Instead, we select only the grayscale
image representation and use this as our malware-image setting.

This setting is important because saliency in a malware image does not necessarily have the same meaning as
saliency in a natural image. In natural images, highly activated regions may correspond to human-interpretable
objects or visual features, while in malware images, such regions may instead reflect byte layout, section structure,
padding, or various artifacts created during the image transformation process. Because of this difference, a CAM-guided
masking strategy that works well for medical images or natural-image classification may not directly transfer to malware
image classification. Therefore, we treat saliency-guided cutout as a method that must be tested within the malware-image
setting rather than assumed to work in the same way across different image domains.

\subsection{Model Scope}

All reported experiments are based on ResNet18~\cite{resnet}. This is a deliberate scope restriction in the current
controlled study.
ResNet18 is a suitable first architecture because it is a standard CNN baseline, trains efficiently,
and is widely used in image-classification experiments. The model registry adapts the classifier head to the number of
classes in each dataset. For small images, the implementation also uses a small-image stem rather than the
larger ImageNet-style stem. Since the current results are ResNet18-only, this chapter does not claim that the
conclusions are architecture independent. Instead, the results should be interpreted as a controlled ResNet18
study that establishes whether saliency-guided cutout is sufficiently promising to justify broader model sweeps.

\section{Methodology}\label{sect:meth}

In this section, we provide relevant details on our experimental design. We also discuss the datasets that we use.

\subsection{Experimental Conditions}

In our experiments, we compare four training scenarios, as summarized in Table~\ref{tab:conditions}. The no-cutout
scenario establishes the baseline for a dataset and the model. The random cutout scenario is the standard cutout regularization
control. The low-saliency and high-saliency scenarios are our proposed CAM-guided variants.

\begin{table}[!htb]
\centering
\caption{Training conditions for cutout experiments}\label{tab:conditions}
\begin{adjustbox}{scale=0.85}
\begin{tabular}{lll}
\toprule
\textbf{Condition} & \textbf{Training data} & \textbf{Cutout region selection} \\
\midrule
No cutout & Original training images & No region is removed \\ \midrule
\multirow{2}{*}{Random cutout} & Original images  & \multirow{2}{*}{Square region selected randomly}\\
                         & plus~$M$ cutout copies &  \\ \midrule
\multirow{3}{*}{Low-saliency cutout} & \multirow{2}{*}{Original images}  & Square region selected from\\
                                 & \multirow{2}{*}{plus~$M$ cutout copies} &  low-saliency candidate windows \\
                                                                          && using teacher HiResCAM map \\ \midrule
\multirow{3}{*}{High-saliency cutout} & \multirow{2}{*}{Original images}  & Square region selected from\\
                                 & \multirow{2}{*}{plus~$M$ cutout copies} &  high-saliency candidate windows \\
                                                                          && using teacher HiResCAM map \\
\bottomrule
\end{tabular}
\end{adjustbox}
\end{table}

Recall from Section~\ref{sect:DCDA} that, for each dataset, the cutout conditions are evaluated with~$M=4$ and~$M=8$.
For both values of~$M$, we experiment with square cutout regions that cover 
approximately 5\%, 10\%, 20\%, or 30\% of the input image.
The augmented training set therefore contains the original image plus four or eight occluded copies, depending on
the value of~$M$. Validation data is never cut out, which is essential for interpreting the result as regularization
rather than as matched-distribution training on a transformed validation set.

\subsection{Pipeline Overview}

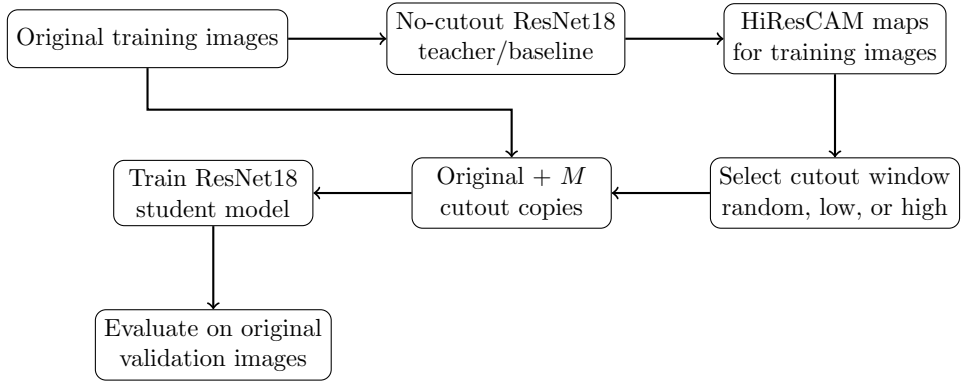
\begin{figure}[!htb]
\centering
\begin{tikzpicture}[node distance=1.45cm, every node/.style={font=\small,scale=0.85}]
\tikzstyle{box}=[draw, rounded corners, align=center, minimum width=3.1cm, minimum height=0.9cm]
\tikzstyle{arrow}=[->, thick]
\node[box] (data) {Original training images};
\node[box, right=1.3cm of data] (teacher) {No-cutout ResNet18\\teacher/baseline};
\node[box, right=1.3cm of teacher] (cam) {HiResCAM maps\\for training images};
\node[box, below=1.1cm of cam] (select) {Select cutout window\\random, low, or high};
\node[box, left=1.3cm of select] (aug) {Original + $M$\\cutout copies};
\node[box, left=1.3cm of aug] (student) {Train ResNet18\\student model};
\node[box, below=1.1cm of student] (eval) {Evaluate on original\\validation images};
\draw[arrow] (data) -- (teacher);
\draw[arrow] (teacher) -- (cam);
\draw[arrow] (cam) -- (select);
\draw[arrow] (select) -- (aug);
\draw[arrow] (aug) -- (student);
\draw[arrow] (student) -- (eval);
\draw[arrow] (data.south) -- ++(0,-0.55) -| (aug.north);
\end{tikzpicture}
\caption{Controlled cutout pipeline}\label{fig:pipeline}
\end{figure}

Figure~\ref{fig:pipeline} illustrates our experimental pipeline. First, a no-cutout ResNet18 model is trained on the original
training images. This model serves two roles: it is the no-cutout baseline, and it provides the teacher checkpoint used
to compute HiResCAM maps for CAM-guided cutout. For the random condition, no teacher is needed. For the low-saliency
and high-saliency conditions, the teacher model generates saliency maps for training images. Candidate square windows
are scored by saliency, and the cutout location is chosen from the low-scoring or high-scoring candidates depending
on the condition. The student model is then trained on the original images plus the selected cutout copies. Finally,
each trained student model is evaluated on the original validation split.

\subsection{CAM-Guided Cutout Implementation}

The saliency-guided cutout implementation is designed to mimic standard cutout as closely as possible. For a given image,
cutout size, and augmentation index, the method selects one square window and fills it with normalized (i.e., black) value.
The random condition selects this window randomly, while the CAM-guided conditions select it using a HiResCAM saliency map.

For CAM-guided cutout, the teacher model is the no-cutout ResNet18 checkpoint trained with the same dataset and seed.
The implementation automatically resolves the final convolutional target layer and computes a HiResCAM map for the
teacher's predicted class. The heatmap is normalized and resized to the input image size. Candidate cutout windows are
then evaluated according to the average saliency inside the square. In the low-saliency condition, the method forms a
candidate set from the lowest-scoring~10\%\ of valid windows. In the high-saliency condition, the candidate set is formed 
from the highest-scoring~10\%. One window is selected from the appropriate candidate set using a specified,
copy-specific seed value. Selecting from a candidate set, rather than always selecting the absolute minimum or
maximum window, preserves diversity among the~$M$ augmented copies.

Because CAM generation is expensive, the implementation includes saliency and window caching. CAM saliency maps
are cached as CPU tensors under a CAM cache directory, and selected window coordinates are cached under a window
subdirectory. This cache is especially important for~$224\times 224$ malware-image datasets, where many candidate windows
must be evaluated. The augmented copies are static: a sample index and augmentation index map to the same selected
window in every epoch. Caching prevents repeated CAM pooling and candidate-window selection across epochs. This makes
the low-saliency and high-saliency runs feasible while preserving the intended experimental design.

\subsection{Repository and Code Organization}

Our code can be found in the public CAMRegularization repository~\cite{GH}. The active workflow is standard training
through \texttt{train.py}, with dataset and model selection handled by registries. The main source files implement training,
evaluation, cutout augmentation, CAM generation, dataset loading, model construction, logging, utility functions, plotting,
and CAM-cutout validation.

The \texttt{train.py} script parses the run configuration, loads the dataset, builds the selected model, optionally loads
the teacher checkpoint for CAM-guided cutout, wraps the training dataset with the cutout augmentation dataset, trains
for the requested number of epochs, and writes the run outputs. The \texttt{engine.py} file contains the epoch-level
training and evaluation routines. The \texttt{cutout.py} file implements the augmented dataset wrapper for random,
low-saliency, and high-saliency cutout. The \texttt{cam\_masking.py} file implements HiResCAM saliency generation.
The \texttt{dataset\_registry.py} and \texttt{model\_registry.py} files isolate dataset loading and model construction
so that the same training pipeline can be used for RawMal-TF and CIFAR-100.

Each committed run folder contains three artifacts: \texttt{config.json}, \texttt{metrics.csv}, and
\texttt{metrics\_plot.png}. The configuration file records the resolved inputs for the run. The metrics file records
per-epoch training loss, training top-1 accuracy, validation loss, validation top-1 accuracy, learning rate, and evaluation
split. The metrics plot visualizes training and validation loss and accuracy across epochs. The results reported in this
chapter are computed from these per-run metrics files and from the repository summary files generated from them.

The current archive contains~168 physical run folders. No-cutout runs are repeated under area directories for matched
folder organization even though cutout area does not apply to the baseline. The summary procedure compares hashes,
selects one baseline observation per dataset and seed, and excludes duplicate copies. After this deduplication,
150 runs are used for analysis: 144 augmented runs and six no-cutout baselines. All expected augmented combinations
are present. One differing duplicate CIFAR-100 baseline copy is excluded, while the repeated modal baseline is retained.
This handling prevents baseline duplication from inflating seed counts or paired comparisons.

\subsection{Datasets}

Our experiments use the two datasets summarized in Table~\ref{tab:datasets}. RawMal-TF is the primary dataset and
provides the malware-family classification setting used to evaluate low- and high-saliency
cutout~\cite{stamp2024malwareimages_arxiv,rawmaltf}. CIFAR-100 is included
as a natural-image comparison dataset so that the cutout behavior can be compared with a standard
computer-vision benchmark~\cite{cifar100}.

\begin{table}[!htb]
\centering
\caption{Datasets analyzed}\label{tab:datasets}
\begin{adjustbox}{scale=0.85}
\begin{tabular}{llll}
\toprule
\textbf{Dataset} & \textbf{Type} & \textbf{Characteristics} & \textbf{Role} \\
\midrule
\multirow{3}{*}{RawMal-TF} & \multirow{3}{*}{Malware images}
& 17 malware families & \multirow{3}{*}{Main empirical focus} \\
&& 1,000 samples per family \\
&& grayscale images  \\ \midrule
\multirow{2}{*}{CIFAR-100} & \multirow{2}{*}{Natural images} & 100 classes & \multirow{2}{*}{Natural-image comparison} \\
&& standard benchmark \\
\bottomrule
\end{tabular}
\end{adjustbox}
\end{table}

For RawMal-TF, the dataset is loaded through the \texttt{drive\_zip} registry entry. The run configuration
sets \texttt{grayscale=true} and uses the include pattern \texttt{\_grayscale.tiff\$}. This ensures that the
RawMal-TF experiment is limited to grayscale images and does not mix the eight available image representations.

\subsection{Training Configuration and Metrics}

Across all reported experiments, we use ResNet18, seed values 42, 43, and~44, 100 epochs, batch size~128,
SGD optimization, learning rate~0.1, momentum~0.9, weight decay~0.0005, a cosine learning-rate scheduler,
five warmup epochs, automatic mixed precision, and deterministic training transforms. Each run uses a validation
split value of~0.1. Validation data remain unmasked. No held-out test results are reported in this chapter because the
committed \texttt{metrics.csv} files contain validation metrics only. For CAM-guided cutout, the teacher model is
ResNet18 and the teacher checkpoint is the no-cutout run for the same dataset and seed. The CAM layer is selected
automatically.

The main reported metric is best validation top-1 accuracy. We report the mean and sample standard deviation across
the three seeds for each value of~$M$, cutout area, and condition combination. We also compute paired
differences within each seed before averaging, so that low-saliency and high-saliency cutout are compared with the
corresponding random cutout run under the same dataset, seed, area, and~$M$.

Several secondary metrics characterize convergence and stability. Final validation accuracy is the value at epoch~100.
Best-to-final degradation measures the decline from the best epoch to epoch~100. Normalized validation area under the
learning curve (AULC) summarizes performance over all 100 epochs. The final-20 standard deviation measures
late-training variation, maximum validation drawdown records the largest decline from a previously achieved running
best, and collapse-event count records entries into a state at least five percentage points below the previous running
best. These secondary metrics help distinguish a condition that produces one favorable epoch from a condition that
performs consistently across training. With only three seeds, confidence intervals and variance estimates are exploratory,
and we avoid strong statistical-significance claims.

\section{Results}\label{sect:res}

In this section, we first present our main experimental results involving the RawMal-TF dataset. Then we give
results for analogous experiments on CIFAR-100. We conclude this section with a cross-dataset summary and
representative learning curves.

\subsection{RawMal-TF Malware Results}\label{subsect:rawmal-results}

RawMal-TF experiments are the main focus of this study. Across the expanded seed and area sweep, saliency-guided
cutout did not improve grayscale malware image classification. The no-cutout baseline obtained the highest mean
best validation accuracy, and every cutout condition produced a lower mean accuracy.

The no-cutout baseline reaches~$72.83\%\pm0.16\%$ mean best validation accuracy. The best cutout result is random
cutout with~$M=4$ and~30\%\ area, which reaches~$71.55\%\pm0.45\%$. The strongest low-saliency result 
is~$71.43\%\pm1.12\%$ for~$M=4$ and~30\%\ area. Thus, the best random cutout condition is~1.28 percentage 
points below the no-cutout baseline, and the best low-saliency condition is~1.39 percentage points below the baseline.

Table~\ref{tab:rawmal} gives all RawMal-TF area and~$M$ combinations. The final two columns give the mean paired
change relative to the matched random cutout control. Positive values favor the CAM-guided condition.

\begin{table}[!htb]
\centering
\caption{RawMal-TF mean best validation accuracy across seeds 42, 43, and 44 (accuracy entries are mean $\pm$ sample standard deviation in percent; paired changes are percentage points; the no-cutout baseline is $72.83\pm0.16$)}
\label{tab:rawmal}
\begin{adjustbox}{scale=0.725}
\begin{tabular}{ccccccc}
\toprule
$\boldsymbol{M}$ & \textbf{Area (\%)} & \textbf{Random} & \textbf{Low saliency} & \textbf{High saliency}
& \textbf{Low--random} & \textbf{High--random} \\
\midrule
4 & 5 & 70.87 $\pm$ 0.48 & 70.94 $\pm$ 0.63 & 70.66 $\pm$ 1.03 & +0.08 & -0.21 \\
4 & 10 & 70.88 $\pm$ 0.66 & 71.02 $\pm$ 0.50 & 70.42 $\pm$ 0.62 & +0.14 & -0.46 \\
4 & 20 & 70.62 $\pm$ 0.90 & 71.08 $\pm$ 0.32 & 69.68 $\pm$ 0.74 & +0.46 & -0.95 \\
4 & 30 & \textbf{71.55 $\pm$ 0.45} & 71.43 $\pm$ 1.12 & 69.54 $\pm$ 0.71 & -0.12 & -2.00 \\ \midrule
8 & 5 & 70.59 $\pm$ 0.43 & 70.02 $\pm$ 0.57 & 70.01 $\pm$ 0.24 & -0.56 & -0.58 \\
8 & 10 & 70.97 $\pm$ 0.81 & 70.63 $\pm$ 0.57 & 69.84 $\pm$ 0.77 & -0.34 & -1.13 \\
8 & 20 & 70.93 $\pm$ 0.33 & 70.77 $\pm$ 1.09 & 69.75 $\pm$ 0.60 & -0.16 & -1.18 \\
8 & 30 & 71.23 $\pm$ 0.62 & 70.83 $\pm$ 0.85 & 69.67 $\pm$ 0.30 & -0.40 & -1.56 \\
\bottomrule
\end{tabular}
\end{adjustbox}
\end{table}

The low-saliency comparison is small and seed-sensitive. Across the eight matched area/$M$ cells, low-saliency
minus random ranges from~$-0.56$ to~$+0.46$ percentage points. One cell is positive for all three seeds, one cell is
negative for all three seeds, and the remaining six cells have mixed signs. Therefore, these experiments do
not support a reliable low-saliency advantage over random cutout on RawMal-TF.

High-saliency cutout is even less favorable. Its mean paired effect relative to random cutout ranges from~$-2.00$ 
to~$-0.21$ percentage points. Six of the eight matched cells are negative for all three seeds, and the other two are mixed.
The largest deficit occurs for~$M=4$ and~30\% area. This is consistent with the possibility that masking the region
most strongly associated with the teacher prediction removes useful malware-family evidence without providing a
compensating regularization benefit.

\begin{figure}[!htb]
\centering
\begin{tikzpicture}[scale=0.90, every node/.style={scale=0.85}]
\begin{groupplot}[
    group style={group size=2 by 1,horizontal sep=1.25cm},
    width=0.47\textwidth,
    height=6.2cm,
    xlabel={Cutout area (\%)},
    ylabel={Best validation error (\%)},
    xtick={5,10,20,30},
    ymin=26.0,ymax=31.5,
    grid=major,
    legend style={legend columns=4,legend cell align=left,nodes={scale=0.95, transform shape}},
]
\nextgroupplot[title={$M=4$},legend to name=rawmalerrorlegend]
\addplot[black,dashed,thick,mark=none] table[col sep=comma,x=area,y=baseline_error]{data/rawmal_error_m4.csv};
\addlegendentry{No cutout}
\addplot[blue,thick,mark=square*,error bars/y dir=both,error bars/y explicit] table[col sep=comma,x=area,y=random_error,y error=random_sd]{data/rawmal_error_m4.csv};
\addlegendentry{Random}
\addplot[green!55!black,thick,mark=triangle*,error bars/y dir=both,error bars/y explicit] table[col sep=comma,x=area,y=low_error,y error=low_sd]{data/rawmal_error_m4.csv};
\addlegendentry{Low saliency}
\addplot[red!75!black,thick,mark=diamond*,error bars/y dir=both,error bars/y explicit] table[col sep=comma,x=area,y=high_error,y error=high_sd]{data/rawmal_error_m4.csv};
\addlegendentry{High saliency}
\nextgroupplot[title={$M=8$},ylabel={}]
\addplot[black,dashed,thick,mark=none] table[col sep=comma,x=area,y=baseline_error]{data/rawmal_error_m8.csv};
\addplot[blue,thick,mark=square*,error bars/y dir=both,error bars/y explicit] table[col sep=comma,x=area,y=random_error,y error=random_sd]{data/rawmal_error_m8.csv};
\addplot[green!55!black,thick,mark=triangle*,error bars/y dir=both,error bars/y explicit] table[col sep=comma,x=area,y=low_error,y error=low_sd]{data/rawmal_error_m8.csv};
\addplot[red!75!black,thick,mark=diamond*,error bars/y dir=both,error bars/y explicit] table[col sep=comma,x=area,y=high_error,y error=high_sd]{data/rawmal_error_m8.csv};
\end{groupplot}
\path (group c1r1.south east) -- node[anchor=north,yshift=-32pt] {\pgfplotslegendfromname{rawmalerrorlegend}} (group c2r1.south west);
\end{tikzpicture}
\caption{RawMal-TF best validation error by cutout area (lower is better; error bars denote sample standard deviation across seeds)}\label{fig:rawmal-error}
\end{figure}
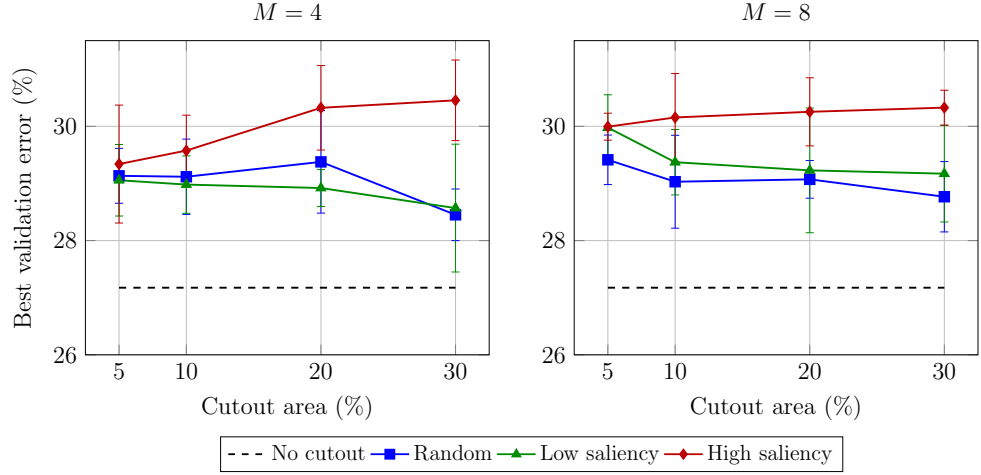

Figure~\ref{fig:rawmal-error} plots validation error rather than validation accuracy, so that differences among conditions
are easier to observe. The no-cutout error is lower than every cutout curve. Low-saliency and random cutout remain close
to one another, while high-saliency error generally increases as the cutout area grows.

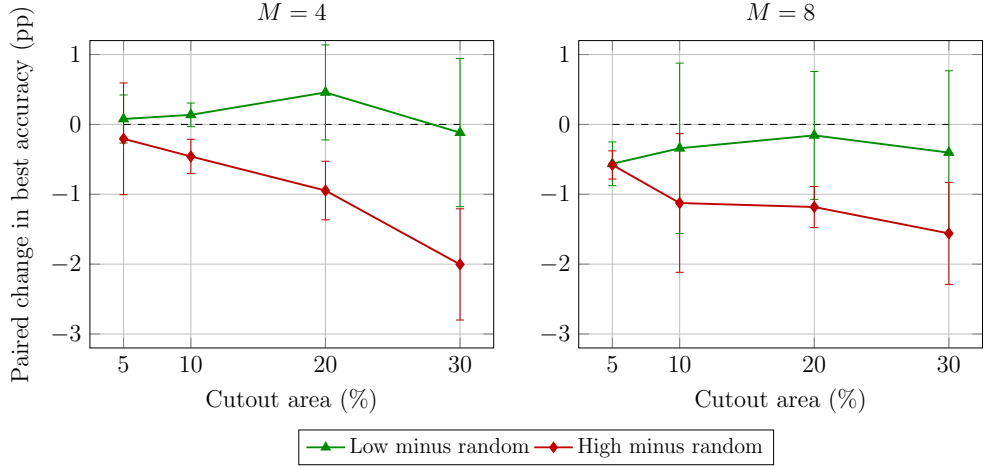
\begin{figure}[!htb]
\centering
\begin{tikzpicture}[scale=0.90, every node/.style={scale=0.85}]
\begin{groupplot}[
    group style={group size=2 by 1,horizontal sep=1.25cm},
    width=0.47\textwidth,
    height=6.1cm,
    xlabel={Cutout area (\%)},
    ylabel={Paired change in best accuracy (pp)},
    xtick={5,10,20,30},
    ymin=-3.2,ymax=1.2,
    grid=major,
    legend style={legend columns=2,legend cell align=left,nodes={scale=0.95, transform shape}},
]
\nextgroupplot[title={$M=4$},legend to name=rawmalpairedlegend]
\addplot[black,dashed,domain=5:30,samples=2,forget plot]{0};
\addplot[green!55!black,thick,mark=triangle*,error bars/y dir=both,error bars/y explicit] table[col sep=comma,x=area,y=low_effect,y error=low_sd]{data/rawmal_paired_m4.csv};
\addlegendentry{Low minus random}
\addplot[red!75!black,thick,mark=diamond*,error bars/y dir=both,error bars/y explicit] table[col sep=comma,x=area,y=high_effect,y error=high_sd]{data/rawmal_paired_m4.csv};
\addlegendentry{High minus random}
\nextgroupplot[title={$M=8$},ylabel={}]
\addplot[black,dashed,domain=5:30,samples=2,forget plot]{0};
\addplot[green!55!black,thick,mark=triangle*,error bars/y dir=both,error bars/y explicit] table[col sep=comma,x=area,y=low_effect,y error=low_sd]{data/rawmal_paired_m8.csv};
\addplot[red!75!black,thick,mark=diamond*,error bars/y dir=both,error bars/y explicit] table[col sep=comma,x=area,y=high_effect,y error=high_sd]{data/rawmal_paired_m8.csv};
\end{groupplot}
\path (group c1r1.south east) -- node[anchor=north,yshift=-32pt] {\pgfplotslegendfromname{rawmalpairedlegend}} (group c2r1.south west);
\end{tikzpicture}
\caption{RawMal-TF paired effects relative to matched random cutout (positive values favor CAM-guided cutout; 
error bars denote sample standard deviation across seeds)}
\label{fig:rawmal-paired}
\end{figure}

Figure~\ref{fig:rawmal-paired} emphasizes the primary control comparison. The low-saliency curve remains near zero,
which reflects the inconsistent and small differences in Table~\ref{tab:rawmal}. The high-saliency curve is mostly below
zero and becomes particularly unfavorable for large masks.

\begin{figure}[!htb]
\centering
\begin{tikzpicture}[scale=0.95, every node/.style={scale=1.0}]
\begin{axis}[
    width=7.5cm,
    height=5.0cm,
    colormap={negzeropos}{color=(red!78!black) color=(white) color=(blue!72!black)},
    colorbar,
    colorbar style={
        ylabel={Percentage points vs random},
        yticklabel style={font=\scriptsize,/pgf/number format/fixed,/pgf/number format/precision=1},
        ylabel style={font=\scriptsize}
    },
    point meta min=-2.1,
    point meta max=2.1,
    xtick={0,1,2,3},
    xticklabels={{Low $M=4$},{High $M=4$},{Low $M=8$},{High $M=8$}},
    xticklabel style={scale=0.80, anchor=east, rotate=60},
    ytick={0,1,2,3},
    yticklabels={5,10,20,30},
    yticklabel style={font=\scriptsize},
    ylabel={Cutout area (\%)},
    ylabel style={font=\small},
    xtick style={draw=none},
    ytick style={draw=none},
    enlargelimits=false,
    axis on top,
    nodes near coords={\pgfmathprintnumber[fixed,precision=2,zerofill]{\pgfplotspointmeta}},
    every node near coord/.append style={font=\scriptsize,fill=white,fill opacity=0.78,text opacity=1,inner sep=1.2pt,anchor=center,text height=1.5ex,text depth=0.25ex},
]
\addplot[matrix plot*,mesh/cols=4,point meta=explicit,draw=white,line width=0.5pt]
table[meta=value] {
    x y value
    0 0  0.077600
    1 0 -0.205967
    2 0 -0.563500
    3 0 -0.579800
    0 1  0.137000
    1 1 -0.458233
    2 1 -0.341600
    3 1 -1.125133
    0 2  0.457767
    1 2 -0.946433
    2 2 -0.156967
    3 2 -1.182300
    0 3 -0.117533
    1 3 -2.003933
    2 3 -0.403367
    3 3 -1.559800
};
\end{axis}
\end{tikzpicture}
\caption{RawMal-TF paired-effect heatmap (blue cells favor saliency-guided cutout, red cells favor matched random cutout, 
and near-white cells indicate little difference)}
\label{fig:rawmal-heatmap}
\end{figure}
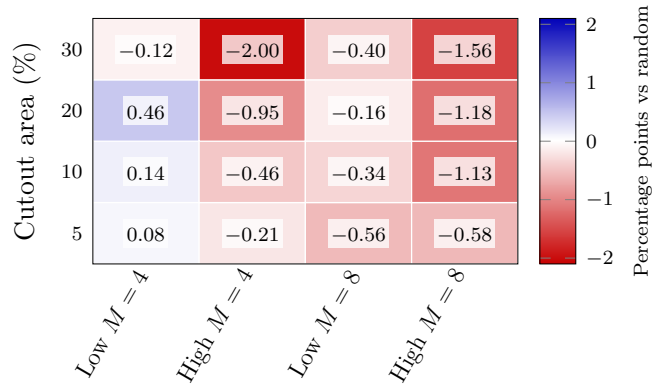

Figure~\ref{fig:rawmal-heatmap} gives a compact view of the same matched comparisons across both augmentation
multiplicities and all four cutout areas. Each cell is the three-seed mean change in best validation accuracy relative
to the random-cutout run with the same seed, area, and value of~$M$. Values near zero dominate the low-saliency
columns, whereas the high-saliency columns are negative throughout and become more unfavorable for larger masks.

\subsection{CIFAR-100 Results}\label{subsect:cifar-results}

CIFAR-100 is included as a natural-image comparison dataset. For this dataset, our no-cutout baseline 
reaches~$62.65\%\pm0.57\%$ for the mean best validation accuracy. Low-saliency cutout with~$M=4$ and!10\%\ area 
is the best CIFAR-100 result at~$63.51\%\pm0.36\%$, which is~0.86 percentage points above the no-cutout baseline 
and~1.25 percentage points above the corresponding random cutout control. Low-saliency cutout with~$M=8$ also
performs well at~5\%\ and~10\%\ area. These CIFAR-100 results are useful because they show that saliency-guided
cutout can have a positive effect in at least one natural image setting, 
which is in contrast to our malware experiments, above.

Table~\ref{tab:cifar} gives the complete CIFAR-100 sweep. Low-saliency cutout is especially useful when the mask
is large enough that random placement becomes destructive. At~20\%\ and~30\%\ area, low-saliency cutout substantially
outperforms matched random cutout for both values of~$M$. However, low-saliency cutout at~30\%\ area remains below
the no-cutout baseline, which shows that ``safer'' placement does not completely eliminate the damage caused by very
large occlusions.

\begin{table}[!htb]
\centering
\caption{CIFAR-100 mean best validation accuracy across seeds 42, 43, and 44 (accuracy entries are mean $\pm$ sample standard deviation in percent; paired changes are percentage points; the no-cutout baseline is $62.65\pm0.57$)}
\label{tab:cifar}
\begin{adjustbox}{scale=0.725}
\begin{tabular}{ccccccc}
\toprule
$\boldsymbol{M}$ & \textbf{Area (\%)} & \textbf{Random} & \textbf{Low saliency} & \textbf{High saliency}
& \textbf{Low--random} & \textbf{High--random} \\
\midrule
4 & 5 & 62.76 $\pm$ 0.46 & 62.88 $\pm$ 0.35 & 60.80 $\pm$ 0.50 & +0.12 & -1.95 \\
4 & 10 & 62.26 $\pm$ 0.34 & \textbf{63.51 $\pm$ 0.36} & 58.85 $\pm$ 0.52 & +1.25 & -3.41 \\
4 & 20 & 59.47 $\pm$ 0.79 & 63.05 $\pm$ 0.35 & 54.07 $\pm$ 0.49 & +3.58 & -5.39 \\
4 & 30 & 53.98 $\pm$ 0.71 & 59.95 $\pm$ 0.73 & 47.26 $\pm$ 0.29 & +5.97 & -6.72 \\ \midrule
8 & 5 & 62.37 $\pm$ 0.38 & 63.22 $\pm$ 0.10 & 60.85 $\pm$ 0.37 & +0.85 & -1.52 \\
8 & 10 & 62.27 $\pm$ 0.16 & 63.31 $\pm$ 0.77 & 58.15 $\pm$ 0.14 & +1.04 & -4.11 \\
8 & 20 & 59.07 $\pm$ 0.50 & 62.75 $\pm$ 0.39 & 51.99 $\pm$ 0.60 & +3.68 & -7.08 \\
8 & 30 & 51.68 $\pm$ 0.86 & 59.59 $\pm$ 0.55 & 45.62 $\pm$ 0.27 & +7.91 & -6.06 \\
\bottomrule
\end{tabular}
\end{adjustbox}
\end{table}

The high-saliency CIFAR-100 result is the opposite of the low-saliency result. High-saliency cutout is worse than
matched random cutout in every area/$M$ cell and for every seed. The paired deficit ranges from~1.52 to~7.08 percentage
points. This is consistent with the intuition that hiding the most important object regions can be counterproductive when
dealing with natural images. The low-saliency result, by contrast, suggests that saliency can identify regions that can be
occluded with less damage to the main object signal.

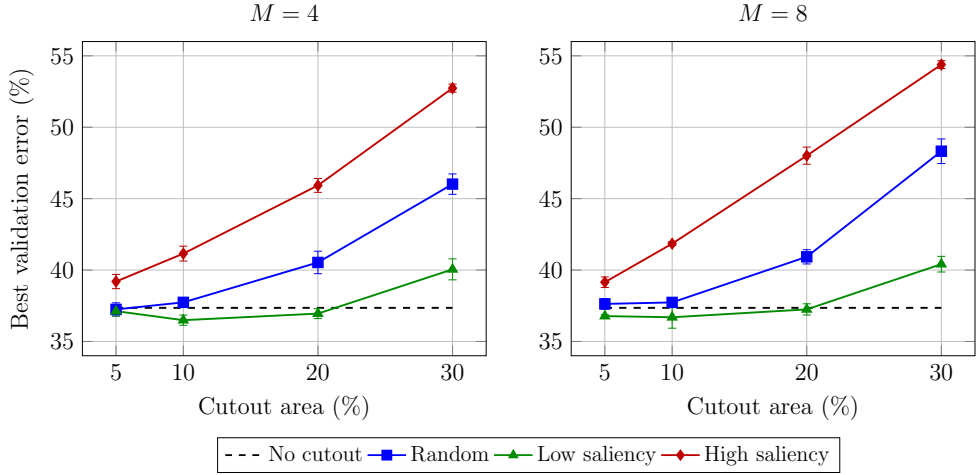
\begin{figure}[!htb]
\centering
\begin{tikzpicture}[scale=0.90, every node/.style={scale=0.85}]
\begin{groupplot}[
    group style={group size=2 by 1,horizontal sep=1.25cm},
    width=0.47\textwidth,
    height=6.2cm,
    xlabel={Cutout area (\%)},
    ylabel={Best validation error (\%)},
    xtick={5,10,20,30},
    ymin=34.0,ymax=56.0,
    grid=major,
    legend style={legend columns=4,legend cell align=left,,nodes={scale=0.95, transform shape}},
]
\nextgroupplot[title={$M=4$},legend to name=cifarerrorlegend]
\addplot[black,dashed,thick,mark=none] table[col sep=comma,x=area,y=baseline_error]{data/cifar_error_m4.csv};
\addlegendentry{No cutout}
\addplot[blue,thick,mark=square*,error bars/y dir=both,error bars/y explicit] table[col sep=comma,x=area,y=random_error,y error=random_sd]{data/cifar_error_m4.csv};
\addlegendentry{Random}
\addplot[green!55!black,thick,mark=triangle*,error bars/y dir=both,error bars/y explicit] table[col sep=comma,x=area,y=low_error,y error=low_sd]{data/cifar_error_m4.csv};
\addlegendentry{Low saliency}
\addplot[red!75!black,thick,mark=diamond*,error bars/y dir=both,error bars/y explicit] table[col sep=comma,x=area,y=high_error,y error=high_sd]{data/cifar_error_m4.csv};
\addlegendentry{High saliency}
\nextgroupplot[title={$M=8$},ylabel={}]
\addplot[black,dashed,thick,mark=none] table[col sep=comma,x=area,y=baseline_error]{data/cifar_error_m8.csv};
\addplot[blue,thick,mark=square*,error bars/y dir=both,error bars/y explicit] table[col sep=comma,x=area,y=random_error,y error=random_sd]{data/cifar_error_m8.csv};
\addplot[green!55!black,thick,mark=triangle*,error bars/y dir=both,error bars/y explicit] table[col sep=comma,x=area,y=low_error,y error=low_sd]{data/cifar_error_m8.csv};
\addplot[red!75!black,thick,mark=diamond*,error bars/y dir=both,error bars/y explicit] table[col sep=comma,x=area,y=high_error,y error=high_sd]{data/cifar_error_m8.csv};
\end{groupplot}
\path (group c1r1.south east) -- node[anchor=north,yshift=-32pt] {\pgfplotslegendfromname{cifarerrorlegend}} (group c2r1.south west);
\end{tikzpicture}
\caption{CIFAR-100 best validation error by cutout area (lower is better; error bars denote sample standard deviation across seeds)}
\label{fig:cifar-error}
\end{figure}

Figure~\ref{fig:cifar-error} uses validation error to make these differences visible. Random cutout and high-saliency
cutout deteriorate rapidly as area increases. Low-saliency cutout is much more robust to area, and its~10\%\ settings
produce the lowest mean error.

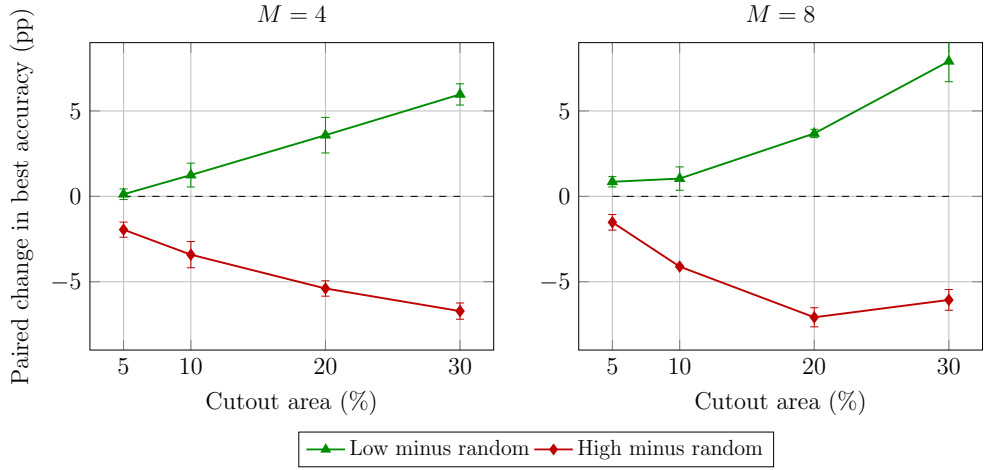
\begin{figure}[!htb]
\centering
\begin{tikzpicture}[scale=0.90, every node/.style={scale=0.85}]
\begin{groupplot}[
    group style={group size=2 by 1,horizontal sep=1.25cm},
    width=0.47\textwidth,
    height=6.1cm,
    xlabel={Cutout area (\%)},
    ylabel={Paired change in best accuracy (pp)},
    xtick={5,10,20,30},
    ymin=-9.0,ymax=9.0,
    grid=major,
    legend style={legend columns=2,legend cell align=left,nodes={scale=0.95, transform shape}},
]
\nextgroupplot[title={$M=4$},legend to name=cifarpairedlegend]
\addplot[black,dashed,domain=5:30,samples=2,forget plot]{0};
\addplot[green!55!black,thick,mark=triangle*,error bars/y dir=both,error bars/y explicit] table[col sep=comma,x=area,y=low_effect,y error=low_sd]{data/cifar_paired_m4.csv};
\addlegendentry{Low minus random}
\addplot[red!75!black,thick,mark=diamond*,error bars/y dir=both,error bars/y explicit] table[col sep=comma,x=area,y=high_effect,y error=high_sd]{data/cifar_paired_m4.csv};
\addlegendentry{High minus random}
\nextgroupplot[title={$M=8$},ylabel={}]
\addplot[black,dashed,domain=5:30,samples=2,forget plot]{0};
\addplot[green!55!black,thick,mark=triangle*,error bars/y dir=both,error bars/y explicit] table[col sep=comma,x=area,y=low_effect,y error=low_sd]{data/cifar_paired_m8.csv};
\addplot[red!75!black,thick,mark=diamond*,error bars/y dir=both,error bars/y explicit] table[col sep=comma,x=area,y=high_effect,y error=high_sd]{data/cifar_paired_m8.csv};
\end{groupplot}
\path (group c1r1.south east) -- node[anchor=north,yshift=-32pt] {\pgfplotslegendfromname{cifarpairedlegend}} (group c2r1.south west);
\end{tikzpicture}
\caption{CIFAR-100 paired effects relative to matched random cutout (positive values favor CAM-guided cutout; 
error bars denote sample standard deviation across seeds)}
\label{fig:cifar-paired}
\end{figure}

Figure~\ref{fig:cifar-paired} presents the same result relative to matched random cutout. The low-saliency advantage
grows as area increases, while high-saliency cutout remains consistently unfavorable. Seven of the eight low-saliency
area/$M$ cells are positive for all three seeds; the~5\%, $M=4$ cell has mixed signs. In contrast, all eight high-saliency 
cells are negative for all three seeds.

\begin{figure}[!htb]
\centering
\begin{tikzpicture}[scale=0.95, every node/.style={scale=1.0}]
\begin{axis}[
    width=7.5cm,
    height=5.0cm,
    colormap={negzeropos}{color=(red!78!black) color=(white) color=(blue!72!black)},
    colorbar,
    colorbar style={
        ylabel={Percentage points vs random},
        yticklabel style={font=\scriptsize,/pgf/number format/fixed,/pgf/number format/precision=1},
        ylabel style={font=\scriptsize}
    },
    point meta min=-8.2,
    point meta max=8.2,
    xtick={0,1,2,3},
    xticklabels={{Low $M=4$},{High $M=4$},{Low $M=8$},{High $M=8$}},
    xticklabel style={scale=0.80, anchor=east, rotate=60},
    ytick={0,1,2,3},
    yticklabels={5,10,20,30},
    yticklabel style={font=\scriptsize},
    ylabel={Cutout area (\%)},
    ylabel style={font=\small},
    xtick style={draw=none},
    ytick style={draw=none},
    enlargelimits=false,
    axis on top,
    nodes near coords={\pgfmathprintnumber[fixed,precision=2,zerofill]{\pgfplotspointmeta}},
    every node near coord/.append style={font=\scriptsize,fill=white,fill opacity=0.78,text opacity=1,inner sep=1.2pt,anchor=center,text height=1.5ex,text depth=0.25ex},
]
\addplot[matrix plot*,mesh/cols=4,point meta=explicit,draw=white,line width=0.5pt]
table[meta=value] {
    x y value
    0 0  0.122933
    1 0 -1.951233
    2 0  0.852867
    3 0 -1.520367
    0 1  1.246933
    1 1 -3.414533
    2 1  1.039733
    3 1 -4.113067
    0 2  3.580367
    1 2 -5.394833
    2 2  3.681867
    3 2 -7.082533
    0 3  5.968500
    1 3 -6.719500
    2 3  7.908267
    3 3 -6.064600
};
\end{axis}
\end{tikzpicture}
\caption{CIFAR-100 paired-effect heatmap (blue cells favor saliency-guided cutout, red cells favor matched random cutout, 
and near-white cells indicate little difference)}
\label{fig:cifar-heatmap}
\end{figure}
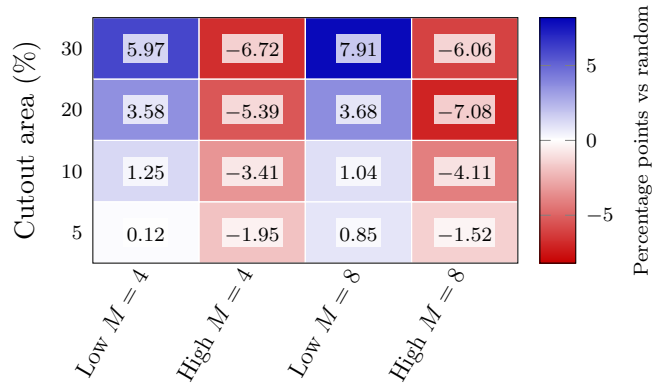

Figure~\ref{fig:cifar-heatmap} summarizes the CIFAR-100 paired effects in the same format. The low-saliency
columns become increasingly positive as the cutout area grows, while every high-saliency cell is negative. This heatmap
makes the interaction between placement rule, mask area, and augmentation multiplicity visible at a glance.

\begin{figure}[!htb]
\centering
\begin{tikzpicture}[scale=0.90, every node/.style={scale=0.85}]
\begin{groupplot}[
    group style={group size=2 by 2,horizontal sep=1.15cm,vertical sep=1.25cm},
    width=0.47\textwidth,
    height=4.65cm,
    xtick={5,10,20,30},
    grid=major,
    legend style={legend columns=4,legend cell align=left,nodes={scale=0.95, transform shape}},
]
\nextgroupplot[title={CIFAR-100, $M=4$},ylabel={Low minus random (pp)},ymin=-1,ymax=10,xticklabels={},legend to name=seedlevellegend]
\addplot[black,dashed,domain=5:30,samples=2,forget plot]{0};
\addplot[blue,thin,mark=square*] table[col sep=comma,x=area,y=seed42]{data/cifar_low_seed_effects_m4.csv};\addlegendentry{Seed 42}
\addplot[orange!85!black,thin,mark=triangle*] table[col sep=comma,x=area,y=seed43]{data/cifar_low_seed_effects_m4.csv};\addlegendentry{Seed 43}
\addplot[purple,thin,mark=diamond*] table[col sep=comma,x=area,y=seed44]{data/cifar_low_seed_effects_m4.csv};\addlegendentry{Seed 44}
\addplot[green!45!black,very thick,mark=*] table[col sep=comma,x=area,y=mean]{data/cifar_low_seed_effects_m4.csv};\addlegendentry{Mean}
\nextgroupplot[title={CIFAR-100, $M=8$},ylabel={},ymin=-1,ymax=10,xticklabels={}]
\addplot[black,dashed,domain=5:30,samples=2,forget plot]{0};
\addplot[blue,thin,mark=square*] table[col sep=comma,x=area,y=seed42]{data/cifar_low_seed_effects_m8.csv};
\addplot[orange!85!black,thin,mark=triangle*] table[col sep=comma,x=area,y=seed43]{data/cifar_low_seed_effects_m8.csv};
\addplot[purple,thin,mark=diamond*] table[col sep=comma,x=area,y=seed44]{data/cifar_low_seed_effects_m8.csv};
\addplot[green!45!black,very thick,mark=*] table[col sep=comma,x=area,y=mean]{data/cifar_low_seed_effects_m8.csv};
\nextgroupplot[title={RawMal-TF, $M=4$},xlabel={Cutout area (\%)},ylabel={Low minus random (pp)},ymin=-2,ymax=1.6]
\addplot[black,dashed,domain=5:30,samples=2,forget plot]{0};
\addplot[blue,thin,mark=square*] table[col sep=comma,x=area,y=seed42]{data/rawmal_low_seed_effects_m4.csv};
\addplot[orange!85!black,thin,mark=triangle*] table[col sep=comma,x=area,y=seed43]{data/rawmal_low_seed_effects_m4.csv};
\addplot[purple,thin,mark=diamond*] table[col sep=comma,x=area,y=seed44]{data/rawmal_low_seed_effects_m4.csv};
\addplot[green!45!black,very thick,mark=*] table[col sep=comma,x=area,y=mean]{data/rawmal_low_seed_effects_m4.csv};
\nextgroupplot[title={RawMal-TF, $M=8$},xlabel={Cutout area (\%)},ylabel={},ymin=-2,ymax=1.6]
\addplot[black,dashed,domain=5:30,samples=2,forget plot]{0};
\addplot[blue,thin,mark=square*] table[col sep=comma,x=area,y=seed42]{data/rawmal_low_seed_effects_m8.csv};
\addplot[orange!85!black,thin,mark=triangle*] table[col sep=comma,x=area,y=seed43]{data/rawmal_low_seed_effects_m8.csv};
\addplot[purple,thin,mark=diamond*] table[col sep=comma,x=area,y=seed44]{data/rawmal_low_seed_effects_m8.csv};
\addplot[green!45!black,very thick,mark=*] table[col sep=comma,x=area,y=mean]{data/rawmal_low_seed_effects_m8.csv};
\end{groupplot}
\path (group c1r2.south east) -- node[anchor=north,yshift=-32pt] {\pgfplotslegendfromname{seedlevellegend}} (group c2r2.south west);
\end{tikzpicture}
\caption{Seed-level paired effects for low-saliency cutout relative to matched random cutout 
(the thick green curve is the three-seed mean)}\label{fig:low-seed-effects}
\end{figure}
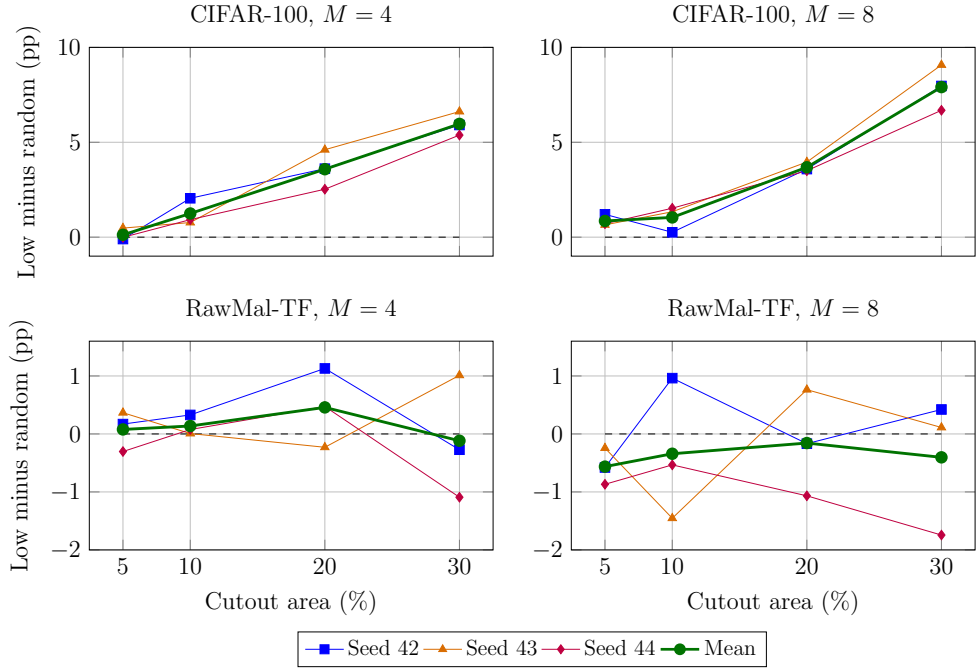

Figure~\ref{fig:low-seed-effects} exposes the individual seed-level low-saliency effects that underlie the means and
error bars. The CIFAR-100 effect is positive and increasingly large for most settings, whereas the RawMal-TF traces
remain close to zero and change sign across seeds. This view makes the contrast between a robust natural-image trend
and a seed-sensitive malware-image result explicit.

\subsection{Cross-Dataset Summary}\label{subsect:cross-summary}

Table~\ref{tab:cross} summarizes the main across-seed findings. The overall pattern is not consistent across image types:
RawMal-TF is best with no cutout, while CIFAR-100 is best with low-saliency cutout at~$M=4$ and~10\%\ area.
The most important difference is not simply the absolute accuracy level, but the relationship between CAM-guided
placement and the matched random control.

\begin{table}[!htb]
\centering
\caption{Cross-dataset summary of best mean validation results}\label{tab:cross}
\begin{adjustbox}{scale=0.85}
\begin{tabular}{lcccc}
\toprule
\multirow{2}{*}{\raisebox{-2pt}{\textbf{Dataset}}} & \multirow{2}{*}{\raisebox{-2pt}{\textbf{No cutout}}} 
	& \multicolumn{3}{c}{\textbf{Best cutout results}}  \\ \cmidrule(lr){3-5}
	& & \textbf{Random} & \textbf{Low saliency} & \textbf{High saliency} \\
\midrule
CIFAR-100 & 62.65 $\pm$ 0.57 & 62.76 $\pm$ 0.46 & \textbf{63.51 $\pm$ 0.36} & 60.85 $\pm$ 0.37 \\
RawMal-TF & \textbf{72.83 $\pm$ 0.16} & 71.55 $\pm$ 0.45 & 71.43 $\pm$ 1.12 & 70.66 $\pm$ 1.03 \\
\bottomrule
\end{tabular}
\end{adjustbox}
\end{table}

\begin{figure}[!htb]
\centering
\begin{tikzpicture}[scale=0.90, every node/.style={scale=0.85}]
\begin{groupplot}[
    group style={group size=2 by 1,horizontal sep=1.35cm},
    width=0.47\textwidth,
    height=6.3cm,
    xtick={1,2,3,4},
    xticklabels={No cutout,Best random,Best low,Best high},
    x tick label style={rotate=25,anchor=east,xshift=2.5pt,scale=0.95},
    y tick label style={scale=0.95},
    ylabel={Best validation accuracy (\%)},
    grid=none,
    tick align=outside,
    major tick length=3pt,
    xtick pos=bottom,
    ytick pos=left,
    axis line style={black},
    title style={scale=0.95},
    error bars/y dir=both,error bars/y explicit,
]
\nextgroupplot[title={CIFAR-100},ybar,bar width=12pt,ymin=59.5,ymax=64.05,ytick={60,61,62,63,64}]
\addplot[fill=blue!55,draw=blue!75!black] table[col sep=comma,x=x,y=accuracy,y error=accuracy_sd]{data/cifar_best_conditions.csv};
\nextgroupplot[title={RawMal-TF},ylabel={},ybar,bar width=12pt,ymin=68.8,ymax=73.125,,ytick={69,70,71,72,73}]
\addplot[fill=green!50!black,draw=green!30!black] table[col sep=comma,x=x,y=accuracy,y error=accuracy_sd]{data/rawmal_best_conditions.csv};
\end{groupplot}
\end{tikzpicture}
\caption{Best mean validation accuracy within each method family. Each cutout bar represents that method's best area and $M$ setting for the corresponding dataset.}\label{fig:cross-best}
\end{figure}
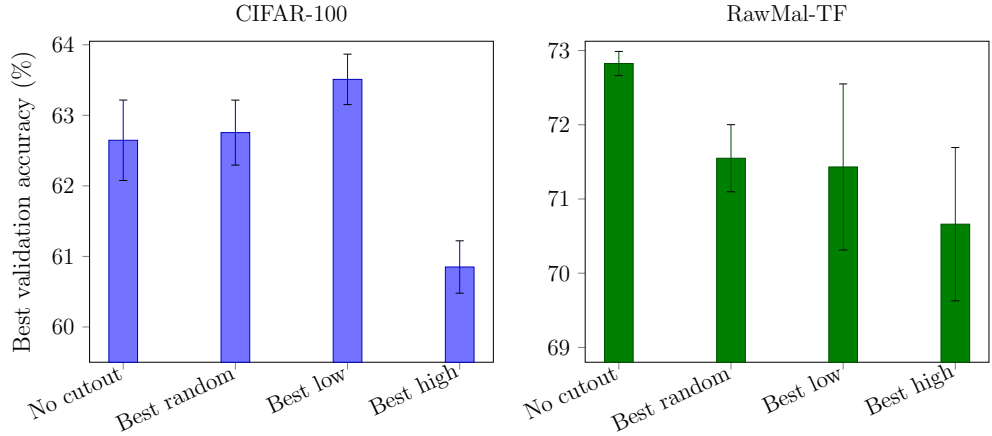

Figure~\ref{fig:cross-best} visualizes the same cross-dataset comparison and makes the different domain-level rankings
immediately visible. Low-saliency cutout supplies the best CIFAR-100 mean, whereas the RawMal-TF no-cutout baseline
remains clearly above the best setting from every cutout family.

\begin{figure}[!htb]
\centering
\begin{tikzpicture}[scale=0.90, every node/.style={scale=0.85}]
\begin{groupplot}[
    group style={group size=2 by 1,horizontal sep=1.35cm},
    width=0.47\textwidth,
    height=6.3cm,
    xtick={1,2,3,4},
    xticklabels={No cutout,Best random,Best low,Best high},
    x tick label style={rotate=25,anchor=east,xshift=2.5pt,scale=0.95},
    y tick label style={scale=0.95},
    grid=none,
    tick align=outside,
    major tick length=3pt,
    xtick pos=bottom,
    ytick pos=left,
    ylabel style={yshift=-5pt},
    axis line style={black},
    title style={scale=0.95},
    legend style={legend columns=2,legend cell align=left,nodes={scale=0.95, transform shape}},
    error bars/y dir=both,error bars/y explicit,
]
\nextgroupplot[title={Whole-curve performance},ylabel={Validation AULC (\%)},ybar=1pt,bar width=6pt,ymin=49,ymax=70,legend to name=secondarymetricslegend]
\addplot[fill=blue!55,draw=blue!75!black] table[col sep=comma,x=x,y=aulc,y error=aulc_sd]{data/cifar_secondary_metrics.csv};
\addlegendentry{CIFAR-100}
\addplot[fill=green!50!black,draw=green!30!black] table[col sep=comma,x=x,y=aulc,y error=aulc_sd]{data/rawmal_secondary_metrics.csv};
\addlegendentry{RawMal-TF}
\nextgroupplot[title={Late-epoch degradation},ylabel={Best-to-final decline (pp)},ybar=1pt,bar width=6pt,ymin=0,ymax=1.8]
\addplot[fill=blue!55,draw=blue!75!black] table[col sep=comma,x=x,y=degradation,y error=degradation_sd]{data/cifar_secondary_metrics.csv};
\addplot[fill=green!50!black,draw=green!30!black] table[col sep=comma,x=x,y=degradation,y error=degradation_sd]{data/rawmal_secondary_metrics.csv};
\end{groupplot}
\path (group c1r1.south east) -- node[anchor=north,yshift=-32pt] {\pgfplotslegendfromname{secondarymetricslegend}} (group c2r1.south west);
\end{tikzpicture}
\caption{Secondary learning-curve metrics for the selected method-family representatives. Error bars denote sample standard deviation across seeds.}\label{fig:secondary-metrics}
\end{figure}
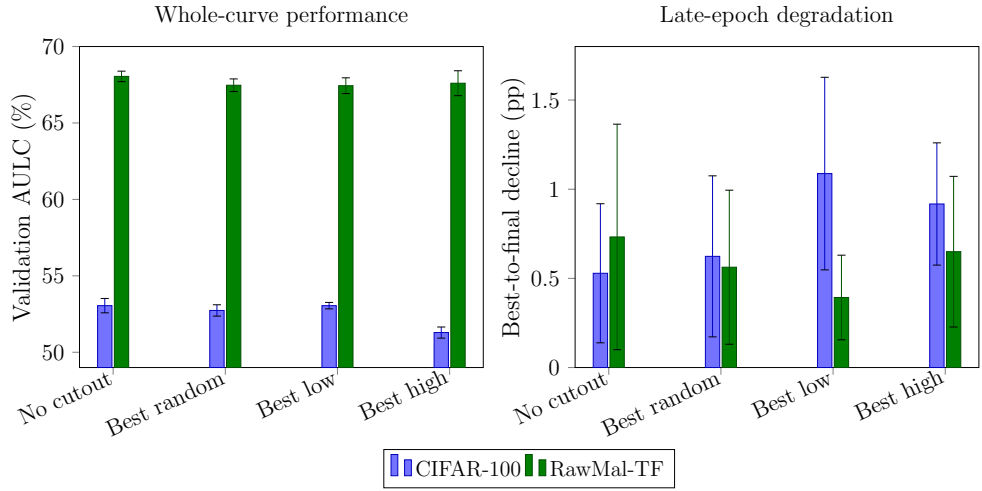

The secondary metrics lead to the same broad conclusion. For RawMal-TF, the no-cutout baseline has a mean normalized
validation AULC of~68.05\%, compared with~67.47\%\ for the best random condition and~67.44\%\ for the best
low-saliency condition. For CIFAR-100, low-saliency cutout can improve peak accuracy without producing an equally
large AULC improvement: the best peak condition, low saliency with~$M=4$ and~10\%\ area, has a mean AULC 
of~53.05\%, the same as for the no-cutout baseline. Low saliency with~$M=8$ and~5\%\ area has the
highest selected AULC at~53.20\%. Thus, the CIFAR-100 advantage is genuine in peak validation accuracy, but the
learning-curve summary indicates that the advantage is concentrated in the later portion of training.
Figure~\ref{fig:secondary-metrics} places these AULC values beside best-to-final degradation for the same selected
conditions. The figure reinforces that peak validation accuracy and whole-curve behavior are related but distinct:
the best CIFAR-100 low-saliency setting improves the peak while remaining close to the baseline in AULC, and none
of the selected RawMal-TF cutout settings exceeds the baseline AULC.

\begin{figure}[!htb]
\centering
\begin{tikzpicture}[scale=0.90, every node/.style={scale=0.85}]
\begin{groupplot}[
    group style={group size=2 by 1,horizontal sep=1.25cm},
    width=0.47\textwidth,
    height=6.0cm,
    xlabel={Epoch},
    xmin=45,xmax=100,
    grid=major,
    legend style={nodes={scale=0.95, transform shape},legend columns=2,
    	legend cell align=left,/tikz/every even column/.append style={column sep=6pt}},
]
\nextgroupplot[title={CIFAR-100},ylabel={Validation error (\%)},ymin=35,ymax=55,legend to name=representativelegend]
\addplot[black,thick] table[col sep=comma,x=epoch,y=cifar_none_val_error]{data/representative_validation_error.csv};
\addlegendentry{No cutout, seed 43}
\addplot[green!55!black,thick] table[col sep=comma,x=epoch,y=cifar_low_val_error]{data/representative_validation_error.csv};
\addlegendentry{Low, $M=4$, 10\%, seed 44}
\addlegendimage{black,thick}
\addlegendentry{No cutout, seed 44}
\addlegendimage{blue,thick}
\addlegendentry{Random, $M=4$, 30\%, seed 43}
\nextgroupplot[title={RawMal-TF},ylabel={},ymin=25,ymax=42]
\addplot[black,thick] table[col sep=comma,x=epoch,y=rawmal_none_val_error]{data/representative_validation_error.csv};
\addplot[blue,thick] table[col sep=comma,x=epoch,y=rawmal_random_val_error]{data/representative_validation_error.csv};
\end{groupplot}
\path (group c1r1.south east) -- node[anchor=north,yshift=-32pt] {\pgfplotslegendfromname{representativelegend}} (group c2r1.south west);
\end{tikzpicture}
\caption{Representative individual-run validation error curves. Aggregate conclusions are based on all three seeds, not on these individual curves.}\label{fig:representative-curves}
\end{figure}
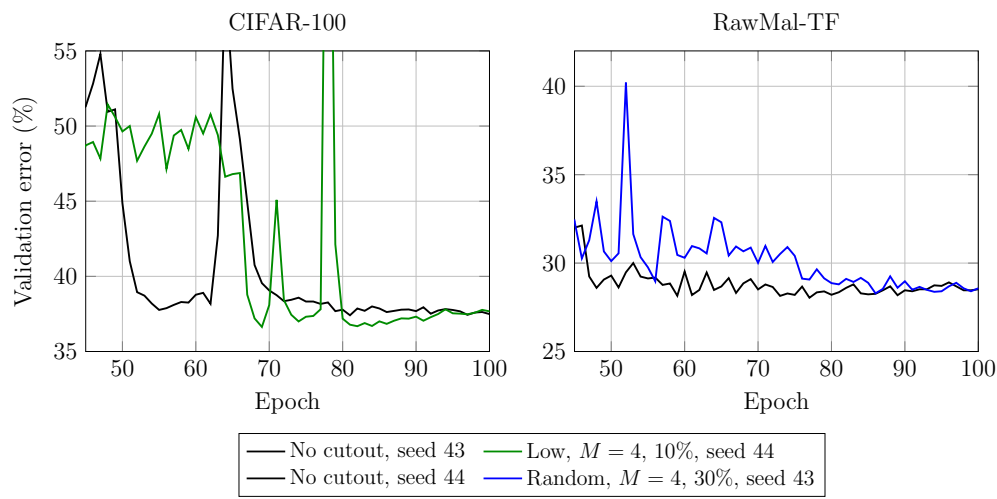

Figure~\ref{fig:representative-curves} presents representative individual run-folder curves using validation error.
The curves are not used in place of the three-seed aggregates; they illustrate
typical late-training behavior. The CIFAR-100 low-saliency condition can reach a lower validation error than the
no-cutout run, while the strongest RawMal-TF random condition remains above the RawMal-TF no-cutout curve.

\section{Discussion}\label{sect:dis}

In this section we expand on the implications of our experimental results. We also consider limitations of the
research presented in this chapter.

\subsection{What the RawMal-TF Result Means}

Our main RawMal-TF result is negative with respect to the original goal of improving malware classification
with CAM-guided regularization. Neither low-saliency nor high-saliency cutout improves over the no-cutout baseline.
More importantly, low-saliency cutout does not consistently improve over the standard random cutout control, and
high-saliency cutout is generally worse. This is the most relevant comparison because random cutout is the direct
control for cutout-style augmentation.

The same basic
pattern is observed across seed values and across cutout areas ranging from~5\%\ through~30\%. The exact ranking of
random and low-saliency cutout changes across some matched cells, but neither produces a mean result that reaches
the no-cutout baseline. The strongest RawMal-TF cutout setting uses random placement, not saliency-guided placement.

One possible explanation is that local occlusion is less appropriate for grayscale malware images than for natural images.
In a natural image, a local square may hide a patch of background, texture, or part of an object. In a malware image, a
local square corresponds to a contiguous region of the transformed binary representation. Depending on the
image-conversion method, such a region may encode structure that is important for family classification. Removing
such structure may not encourage robust visual reasoning---it may simply damage a useful representation.

Another possible explanation is that saliency maps may not identify regions that are useful for cutout in the malware
setting. A high-saliency region in a malware image may correspond to byte layout, packed content, padding, or other
low-level artifacts, and a low-saliency region may still contain structural information needed by the classifier. Conversely,
a low-saliency square may overlap zero padding or an already uniform region and therefore create only a weak
augmentation. The metrics considered in this chapter do not account for zero-padding overlap or 
provide effective perturbation measurements, so these possibilities remain hypotheses for future research.

\subsection{Low-Saliency vs High-Saliency Cutout}

Low-saliency and high-saliency cutout answer different questions. Low-saliency cutout asks whether the model benefits
from occluding regions that the teacher considers less important. High-saliency cutout asks whether the model benefits
from being forced to ``look away'' from the most important regions. Our results suggest that high-saliency cutout
is the riskier strategy, as it is consistently worse than low-saliency cutout on CIFAR-100 and generally worse on RawMal-TF.

The CIFAR-100 result is especially informative. Low-saliency cutout at~10\%\ area improves over no cutout,
while high-saliency cutout sharply reduces performance. Low-saliency placement also protects performance when
random masks become too large. This supports the idea that low-saliency regions can provide safer occlusion locations.
However, the RawMal-TF result shows that this intuition does not automatically transfer to malware images. In the
malware setting, even the best low-saliency condition is worse than no cutout, and its comparison with random cutout
is seed-sensitive.

High-saliency cutout may also create a form of label inconsistency. The augmented image retains the original class label,
but a large square intentionally removes the evidence the teacher most strongly associates with its prediction. On a
natural image this can remove the object itself; on a malware image it can remove a discriminative byte region. Increasing
the number or area of such masks therefore need not behave like beneficial regularization.

\subsection{Effect of Cutout Area and Number of Copies}

The area sweep reveals a strong domain difference. On CIFAR-100, random and high-saliency cutout deteriorate
rapidly as area increases, while low-saliency placement is substantially more robust. At~20\%\ and~30\%\ area, the
low-saliency advantage over random cutout becomes relatively large because saliency guidance avoids the most destructive
locations. Nevertheless, 30\%\ low-saliency masks still underperform the no-cutout baseline.

On RawMal-TF, changing the area does not produce a monotonic effect. The best random and low-saliency means
occur at~30\%\ area with~$M=4$, but both remain below no cutout. The low-minus-random effects remain small relative
to their across-seed variability. Therefore, the area sweep does not reveal a RawMal-TF setting in which
saliency guidance becomes reliably beneficial.

The comparison between~$M=4$ and~$M=8$ is descriptive rather than compute-matched. Each original image yields
five training examples when~$M=4$ and nine when~$M=8$, so the number of augmented examples and optimizer updates
changes with~$M$. On RawMal-TF, the strongest cutout settings use~$M=4$. On CIFAR-100, both values can be useful
for low-saliency placement, but~$M=8$ does not produce a universally better result. Overall, this indicates that~$M$ 
should be treated as a hyperparameter rather than as a universally optimal setting.

\subsection{Structure-Aware Masking for Malware Images}

The negative RawMal-TF result suggests that a masking strategy designed for natural-image saliency may not exploit
the structure of executable binaries. Malware images originate from byte sequences and executable formats, not from
photographic scenes. A square selected only by image saliency ignores information such as file offsets, binary regions,
entropy transitions, executable headers, and Portable Executable (PE) section boundaries.

A more appropriate malware-specific augmentation could align masks with known binary structure. For example, a
structure-aware method could restrict or stratify masks by PE section, distinguish headers from code and data regions,
avoid zero-padded areas, or mask contiguous byte ranges before image conversion. Another approach could combine
saliency with structural constraints, such as selecting a low- or high-saliency region only within a specified PE section.
These methods would preserve the controlled comparison with random masking while testing whether malware-specific
information is more useful than purely image-based square saliency.

The present results should not be interpreted as showing
that all intelligent masking is ineffective for malware. Instead, our results simply show that the 
current low- and high-saliency square strategies, transferred from natural-image reasoning, 
do not improve grayscale RawMal-TF classification.

\subsection{Limitations and Future Work}\label{sect:lim}

The current experiments have several limitations. First, three seeds provide a substantially stronger basis than a
single run, but the sample size remains small. Means, standard deviations, and $t$-based confidence intervals should be
interpreted as exploratory. Additional seeds would be useful for small RawMal-TF paired effects.

Second, the current results are ResNet18-specific. This is useful for a controlled comparison, but it is conceivable that
the same pattern may not hold for DenseNet, EfficientNet, ConvNeXt, vision transformers (ViT), Swin Transformers,
or other architectures~\cite{vit,densenet,swin,convnext,efficientnet}. A broader study should repeat the same controlled
comparison across multiple models.

Third, the current RawMal-TF experiment uses only grayscale images. Grayscale images are most common in malware
analysis, and this restriction provides a reasonable first test case. However, future work should also consider different
image types, while studying saliency within one representation at a time.

Fourth, the current archive supports validation accuracy, validation loss, and learning-curve stability summaries, but
it does not contain held-out test accuracy, macro-F1, per-family metrics, confusion matrices, calibration, or sample-level
predictions. These outputs should be added before making strong test-set or family-specific claims. The current chapter
therefore reports only what is supported by the specific metrics considered.

Fifth, the same seed controls the data split and other sources of training randomness. Within-seed comparisons remain
matched, but across-seed variability combines split sensitivity and optimization sensitivity. A future implementation
should separate the split seed from model initialization, shuffling, and cutout placement.

Sixth, the current teacher generates HiResCAM for its predicted class using the final convolutional layer. Future work
should compare predicted-class and ground-truth targets, HiResCAM and Grad-CAM, and earlier versus later target layers.
Candidate-window percentages should also be varied.

Finally, the present~$M=4$ and~$M=8$ experiments are not compute-matched with no cutout or with each other.
Future work should report wall-clock time, GPU time, and optimizer steps, and should include controls with an equal
training budget. The present chapter should therefore be interpreted as a controlled augmentation sweep rather than
as a training-efficiency study.

\section{Conclusion}\label{sect:con}

This study evaluated saliency-guided cutout regularization for image-based malware classification. The
controlled experiments compared no cutout, standard random cutout, low-saliency cutout, and high-saliency cutout
using ResNet18. For each cutout method, the training set contains the original images plus~$M$ augmented cutout copies,
with~$M=4$ and~$M=8$ tested. The main malware experiment used grayscale RawMal-TF images and
every condition was evaluated with seeds~42, 43, and~44 and cutout areas of~5\%, 10\%,
20\%, and~30\%, with~100 training epochs per run.

The RawMal-TF results show that saliency-guided cutout did not improve malware classification under this setup.
The no-cutout baseline attained~$72.83\%\pm0.16\%$ mean best validation accuracy, while the best cutout condition,
random cutout with~$M=4$ and~30\% area, reached~$71.55\%\pm0.45\%$. Low-saliency cutout was sometimes slightly
better and sometimes slightly worse than matched random cutout, and no low-saliency condition exceeded the no-cutout
baseline. High-saliency cutout was generally harmful.

CIFAR-100 gave a different result. Low-saliency cutout with~$M=4$ and~10\%\ area 
reached~$63.51\%\pm0.36\%$, compared with~$62.65\%\pm0.57\%$ for no cutout.
Low-saliency placement also substantially outperformed random placement for large masks, 
while high-saliency cutout was consistently harmful. This contrast shows that the implementation 
can produce a positive saliency-guided result, but that the benefit does not transfer
to the grayscale malware-image setting.

Overall, our results suggest that saliency-guided cutout is domain dependent. The method may be useful for
natural images, but it does not currently improve grayscale RawMal-TF malware classification. This result
is useful because it clarifies that malware image saliency should not be treated as equivalent to object saliency
in natural images. For malware images, future masking strategies should incorporate executable structure, such as
binary regions or PE-section boundaries, rather than relying only on image saliency and square masks.

\section{Data and Code Availability}

The code and experimental results analyzed in this chapter are available in the public CAMRegularization
GitHub repository at~\cite{GH}. The repository for this chapter is commit
\texttt{cd7f42fe9b3c3b52e291acf795e2dcc7e75bd14f}. The current run folders used for this research contain the
resolved configuration, per-epoch metrics, and training plot for each ResNet18 run. Summary figures and tables are
generated from the current \texttt{metrics.csv} files and repository summary CSV files. 

\bibliographystyle{plain}
\bibliography{references}

\end{document}